# AN OPTIMIZED DENSITY-BASED LANE KEEPING SYSTEM FOR A COST-EFFICIENT AUTONOMOUS VEHICLE PLATFORM: AurigaBot V1

Farbod Younesi, Milad Rabiei*, Soroush Keivanfard*, Mohsen Sharifi*, Marzieh Ghayour Najafabadi, Bahar Moadeli, Arshia Jafari, Mohammad Hossein Moaiyeri**

*Faculty of Electrical Engineering, Shahid Beheshti University, Tehran, Iran*

---

**Abstract**

The development of self-driving cars has garnered significant attention from researchers, universities, and industries worldwide. Autonomous vehicles integrate numerous subsystems, including lane tracking, object detection, and vehicle control, which require thorough testing and validation. Scaled-down vehicles offer a cost-effective and accessible platform for experimentation, providing researchers with opportunities to optimize algorithms under constraints of limited computational power. This paper presents a four-wheeled autonomous vehicle platform designed to facilitate research and prototyping in autonomous driving. Key contributions include (1) a novel density-based clustering approach utilizing histogram statistics for landmark tracking, (2) a lateral controller, and (3) the integration of these innovations into a cohesive platform. Additionally, the paper explores object detection through systematic dataset augmentation and introduces an autonomous parking procedure. The results demonstrate the platform's effectiveness in achieving reliable lane tracking under varying lighting conditions, smooth trajectory following, and consistent object detection performance. Though developed for small-scale vehicles, these modular solutions are adaptable for full-scale autonomous systems, offering a versatile and cost-efficient framework for advancing research and industry applications.

---

* Same Contribution, arranged alphabetically
** Corresponding Author (email: h_moaiyeri@sbu.ac.ir)

## 1. Introduction

Road collisions have always been one of the leading causes of death and injury worldwide. World Health Organization (WHO) reported that approximately 1.19 million people die yearly from road accidents. More than half of road traffic deaths are among vulnerable road users, pedestrians, cyclists, and motorcycles. Also, many road accidents are due to human errors (such as driver distractions or being under the influence of alcohol and other substances) [1] and can be reduced through automation.The latest technologies involving Advanced Driving Assistance Systems (ADAS) and Autonomous Guided Vehicles (AGV) have proven convenient and advantageous in various circumstances. Thus, self-driving cars provide an opportunity for safer roads and a significant yearly reduction in casualties. Furthermore, increasing the number of autonomous vehicles in a society can save many people's time daily. Research also suggests that autonomous driving can reduce the Value of Travel Time Savings (VTTS) by a large margin with privately owned and shared AVs [2].

Therefore, in recent years, the development of self-driving cars has been the main focus of many researchers, universities, and companies worldwide. An autonomous vehicle combines many working units, and many algorithms have been proposed for lane tracking, object detection, car control, etc. These need to be tested on smaller scales to be proven functional. Since scaled-down vehicles are more accessible and cheaper than full-sized vehicles, they provide more experiment, inquiry, and research opportunities. Each year, many competitions are held globally regarding small-scale autonomous vehicles, inspiring researchers and enthusiasts. It is noteworthy that compared to full-scale cars, small-scale robots provide less computational power, so developers should optimize their algorithms to achieve the best results with the lowest possible power consumption.

This paper focuses on presenting a four-wheeled autonomous vehicle platform. Two landmark tracking and lateral control architectures are introduced, perfected for the proposed platform, and implemented. Further, object detection with systematic dataset architecture and autonomous parking procedures are discussed. Incorporating the stated modules implemented on dedicated hardware builds the proposed platform.

The main contributions of this paper are as follows:

- A novel density-based clustering centered on histogram statistics.
- Design of a lateral controller
- Proposition of a platform incorporating the mentioned ideas

The remainder of this paper is arranged in order of the platform's building. Section 2 describes the novel architectures and algorithms. In section 3, the results of the proposed are presented, alongside additional remarks on parking and object detection implemented on the platform. The paper concludes in section 4.

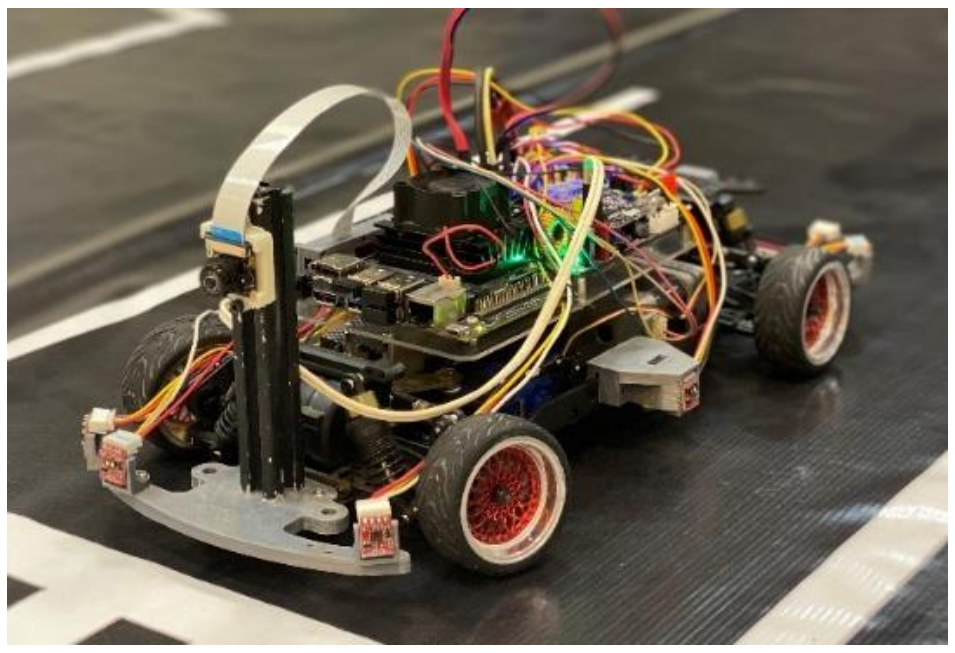

Fig. 1. The 1:10 scale autonomous robot.

## 2. Proposed Methods

Creating a reliable lane-keeping system for self-driving cars and assisted driving involves using a smart controller, which needs ample information from the surroundings to guide the system effectively. The data helps the controller with real-time feedback, ensuring the car navigates well in different driving situations. A strong connection between the controller and extracted data from the surroundings is crucial for making accurate and safe automated driving systems.

A visual perception system capturing and processing road curvature, which provides effective feedback, is needed to enable a suggested controller. A measure of the accuracy of an image-based decision system is based on the perfection of the segmentation technique [3]. This paper suggests an area-oriented unsupervised segmentation algorithm to efficiently and adequately capture and distinguish road lanes and boundaries. Moreover, the extracted data from the segmentation algorithm is further fed into a lateral controller, which is developed to adjust the vehicle's position within the lane continuously, thus enhancing accuracy and maintaining consistent lane-keeping performance.

## 2.1. Landmark tracking system

Feature extraction techniques help identify intrinsic geometric patterns in degraded images, improving analysis and reconstruction despite visual deterioration [4]. Clustering landmarks has numerous applications in spatial databases; Area-oriented image segmentation focuses more on the spatial properties of image pixels. DBSCAN [5][6] (Density-based spatial clustering of applications with noise) is a clustering algorithm that can group arbitrary-shaped unlabeled data points based on their local distances and local density in a space region. This algorithm is especially robust regarding clustering outliers and minimal requirements for the space [ref sigmoid2015]. However, this solution brings up two possible problems:

1. Any data point, whether core or noise point, is assigned to a class.
2. The algorithm may have a huge computational overhead, resulting in processing delays in relatively small edge devices.

We propose a method for detecting, classifying, and tracking landmarks for our specific challenge. In the domain of mobile vehicles, even in extensively filtered images, some noise is redundant but recognized by an uninformed segmentation agent. Following the basic principles of density-based clustering, yet alternating some requirements relative to our domain, we devised an algorithm that ignores possible inevitable noise and is not computationally expensive for our problem, locating and tracking landmarks in a binary 2D image.

### 2.1.1. Locating base points for our landmarks' overall shape objects

Our method begins by identifying key reference points, referred to as core points, that represent the central and distinctive features of each landmark shape. These core points are detected using a convolution-based pattern recognition technique to locate the shape's geometric center or pivotal regions. This step is the foundation for subsequent tracking and classification, ensuring accurate localization of the landmark's overall structure.

For our problem statement, by performing a convolution of a predefined filter with the histogram of the lower third of the binary image, we partition the result into left and right lane regions separated by the highest convolution peak and determine the left-base and right-base points from the respective divisions of the histogram [7][8].

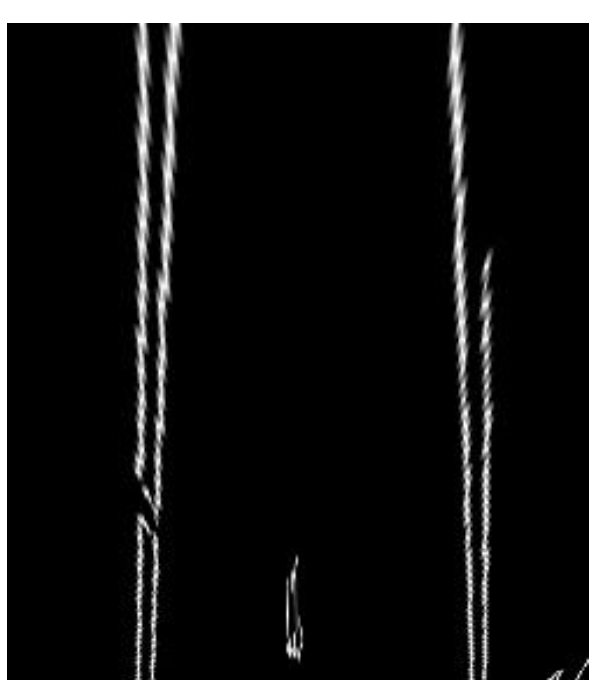

A) Filtered binary image

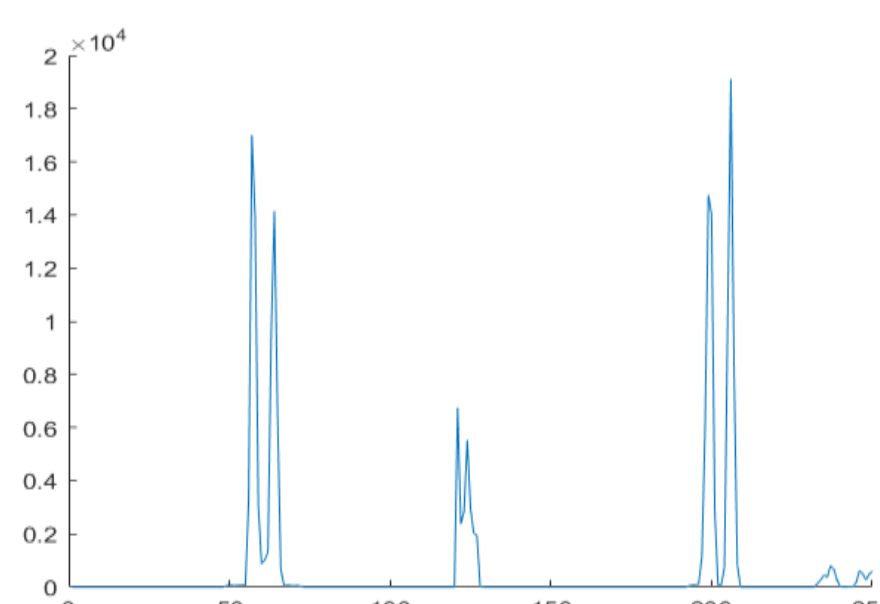

B) Histogram of white pixels on x axis

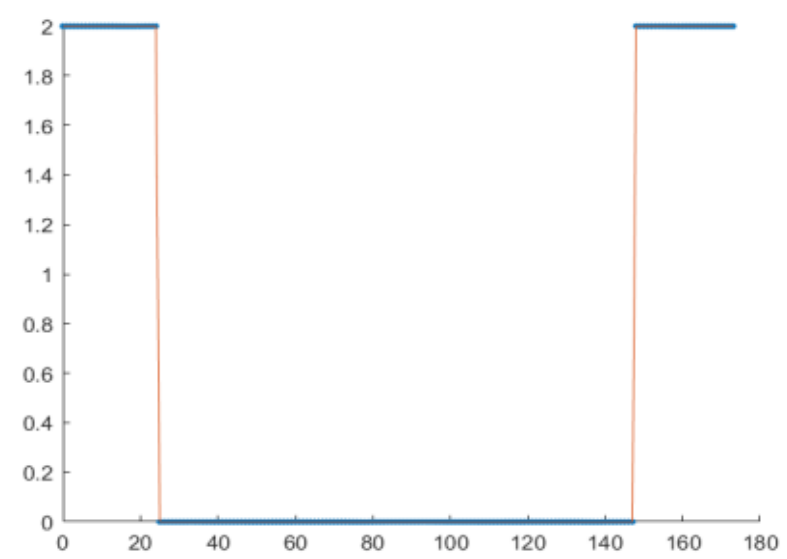

C) Filter applied on histogram

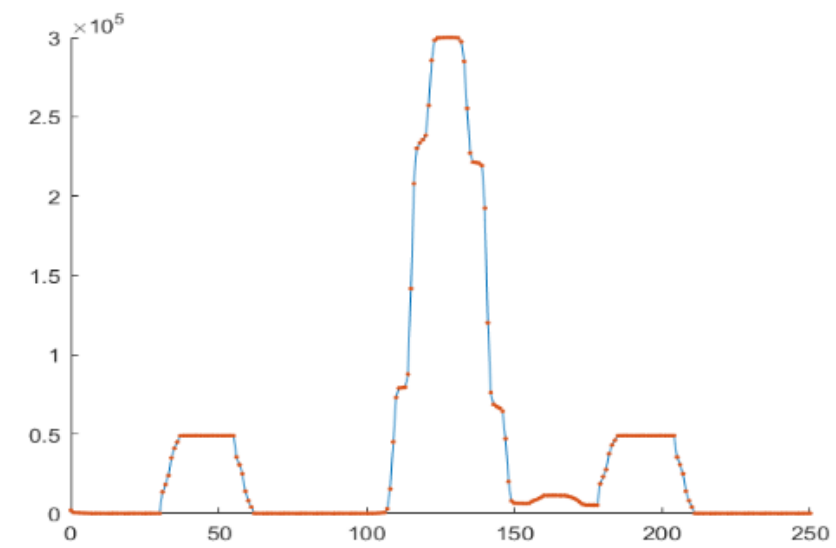

D) Filter convolved on the histogram

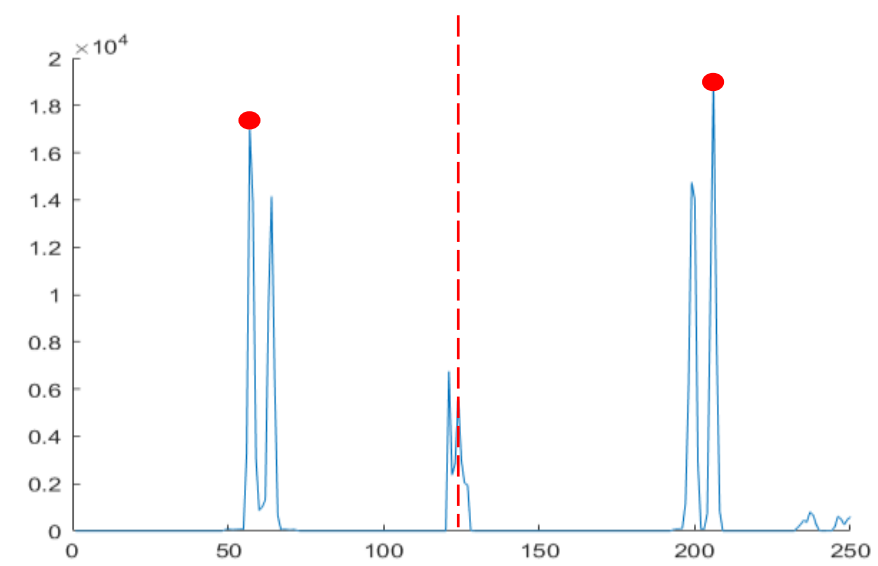

E) Separated histogram, left and right base points

Fig. 2. Applying lane detection on image histogram.

### 2.1.2. Following and classifying the shapes' landmarks

An optimized landmark clustering approach is to use a round ribbon on the margin of a square at the center of the ribbon. The ribbon is responsible for detecting high-density points for the next iteration square placement, where the square captures the pixels neighbor to that point which are in the square's domain and then assigns them to the same class, keeping the characteristics of DBSCAN while improving the execution time.

Unlike the original sliding windows approach for self-driving car prototypes, which determines the next position of the next windows by calculating the mean of white pixels only through the x-axis, in this method, the next position of the square is calculated by the mean of white pixels in the ribbon area around the selected region on both x-axis and y-axis. It allows the capture and marking of lane pixels through relatively sharper turns along the way, which estimates data more accurately and eventually enhances the robot's steering. However, a problem occurs when the agent has to choose between two equally-considerable paths, which raises the importance of the road's heading direction as a factor of preference. An elliptical ribbon searcher agent prefers to track white pixels ahead instead of choosing left or right when following lanes instead of a circular agent. An even more proficient agent would prioritize the front of the agent to its back. This action is performed by defining the agent as a pear-shaped oval due to the non-equal and more disposed distribution of the prominent white pixel density to the front (figure 3).

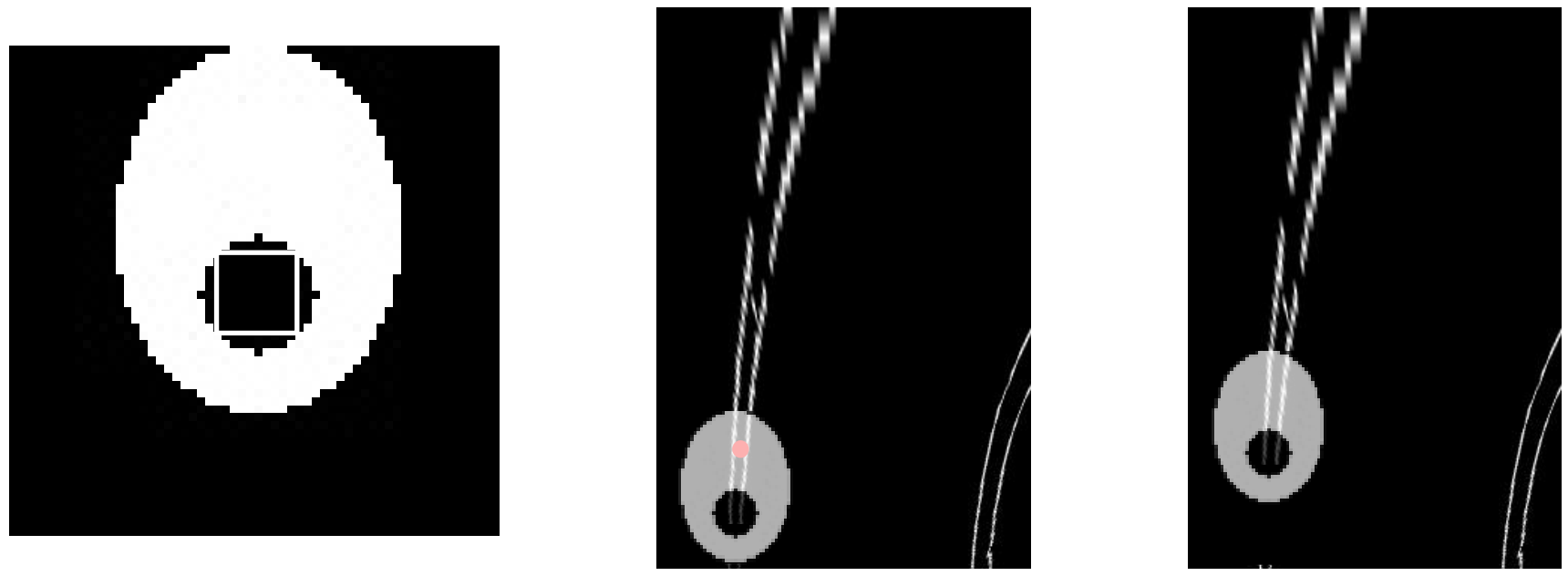

A) Elliptical searcher with the square inside it

B) Calculated density of the white pixels

C) Next position of the agent

Fig. 3. Novel density-based tracking method.

Then, similar to the first method, left and right lanes are determined after white lane pixels are distinguished by being placed inside the square; their positions are recorded and used to approximate the left and right curve polynomials to return the ideal path polynomial eventually. Since this ideal path is calculated by averaging the left and right polynomials, if one is missing in the received image, the ideal path imitates the curve of the available lane with a biased translation.

## 2.2. Lateral controller design

The designed lateral controller operates within a moving Cartesian coordinate system that adjusts its origin based on the road's layout. In this system, the x-axis is oriented in the direction of the road's heading and tangent to the road curve at each moment, while the y-axis is perpendicular to the road. Assuming $(x_0, y_0)$ serves as the system's origin, $x_0$ consistently aligns with the car's camera position, and $y_0$ corresponds to the ideal path between the lane lines.

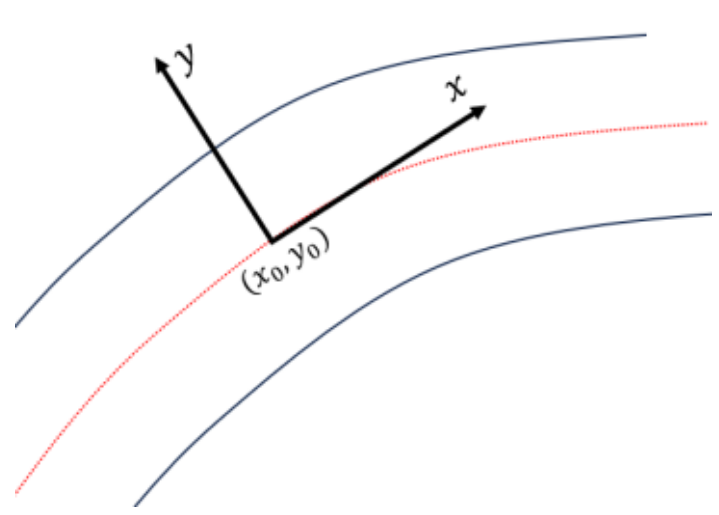

Fig. 4. The coordinate system placement on the road

A planar moving robot model has three degrees of freedom: two for position and one for orientation. As previously noted, $x_0$ is consistently aligned with the car's camera position, making the robot's x-coordinate always zero in this moving Cartesian coordinate system. Consequently, the robot's movement can be effectively described using two variables: $y_R$, which denotes the lateral position relative to this coordinate system, and $\theta$, representing the robot's heading angle. Variable "$y_R$" signifies the disparity between the robot's present position on the road and the lane's center, while $\theta$ indicates the variance in heading between the robot and the road's orientation or x-axis direction. Velocity (v) and steering angle ($\delta$) are the inputs to the robot, yet given our exclusive focus on the lateral controller, $\delta$ stands as the sole output of the controller.

The estimated polynomial (n.m.p.) shows the average trajectory between the two road lane curves. Therefore, the subtraction of the bottommost point of the curve from the center of the frame - which also indicates the robot's center considering the camera's placement on it - results in the "distance error". It determines how far the robot is from the midline at each moment and can be considered as feedback for the variable $y_R$.

The car may also have some angular difference from the road's path heading, and this angular difference with the x-axis gives us the next variable, the robot heading (θ). It can be the difference between the robot's pointing angle from a straight path or the road's curve towards the right or the left turn. This angular difference(α) eventually contributes to the loss of the visible lane by increasing the "distance error" if it is not handled properly, leading to instability and loss of the estimated ideal trajectory; thus, it should be adjusted accordingly with high accuracy and speed. α is the derivative of the bottommost point of the polynomial estimated in degrees.

An IMU offers additional feedback for θ, with a rate distinct from the prior feedback (α). Integrating these two feedback sources establishes a multi-rate sampling system, enhancing the precision of θ estimation and reinforcing stability. This refinement aids in improving the resulting data, thereby pursuing a smoother trajectory.

An escalation in "distance error" almost always results in a corresponding increase in "angle error," and likewise, an elevation in "angle error" leads to an increase in "distance error". Hence, the intended lateral controller must incorporate both feedbacks. This is achieved by defining two distinct controller terms (each derived from a specific state), which are multiplied by appropriate gains and summed together.

Based on prior assumptions, the optimal trajectory entails precisely following the midline. Consequently, adhering to the predefined coordination system, the reference state variable y remains consistently at 0, aiming for zero "distance error" continuously. Thus, the controller includes a term formed by multiplying "distance error" with a gain factor $K_{distance}$.

Additionally, the robot might exhibit steady-state error due to various factors, including imprecise actuators or simplifications made during the coordination assumption process. Consequently, an integrator term is incorporated into the controller, utilizing the feedback from "distance error."

Furthermore, the optimal drive scenario stipulates that the car's heading perfectly aligns with the road heading, aiming for an "angle error" value of 0. Following this procedure, "angle error" is measured in degrees. To convert it into a scalar value, a tangent function is applied to the error in radians, which, when multiplied by a gain factor $K_{angle}$, constitutes another term of the controller. When the robot's speed is considered in $K_{angle}$, it could nearly act as a derivative to the distance error, demonstrating the rate of distance in approaching or moving away from the middle of the road. Ultimately, these three terms are aggregated and utilized as the steering input δ for the system. The final control loop is illustrated in Figure 5. [9][10]

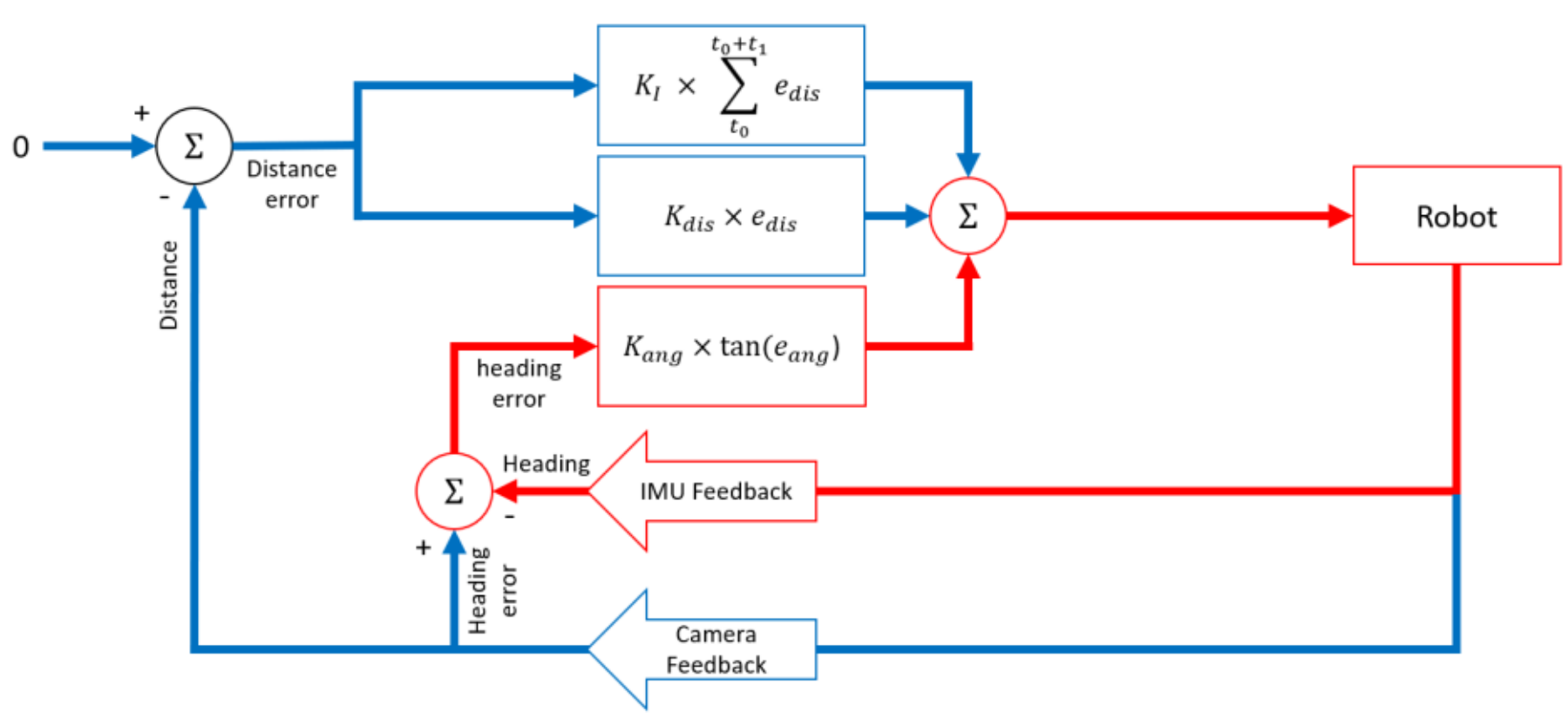

Fig. 5. The diagram of multi-rate control system

## 3. Results, Applications, and the Platform

The proposed methods of vehicle maneuvering have been implemented into a vehicle robot platform, which is discussed later. Results of each are presented in this section in detail, along with several other remarks for a complete vehicle platform, i.e., object detection and vehicle parking.

### 3.1. Lane tracking

Lane detection involves camera calibration, filtering, and detection and tracking. The preprocess is detailed in Figure 6. The road region of the filtered image is specified to be warped to a bird's view, which gives a better sight of the curves. It enables more accurate analysis of the path by generating non-convergent road lanes.

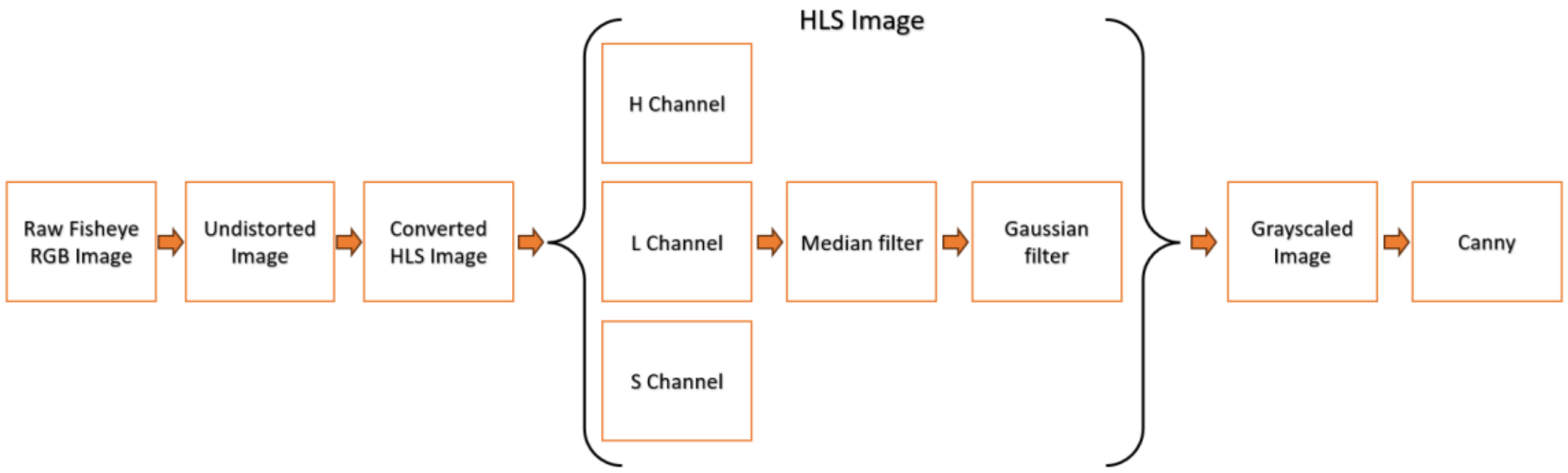

Fig. 6. Preprocess stages

Road lane detection and tracking is performed using the proposed segmentation technique. The ideal path polynomial is determined by extracting right and left lane pixels and averaging the approximated curves of each. Besides efficiency and adequacy, this setup outperforms traditional sliding windows with rectangular divisions since even lane pixels of very sharp curves can be tracked.

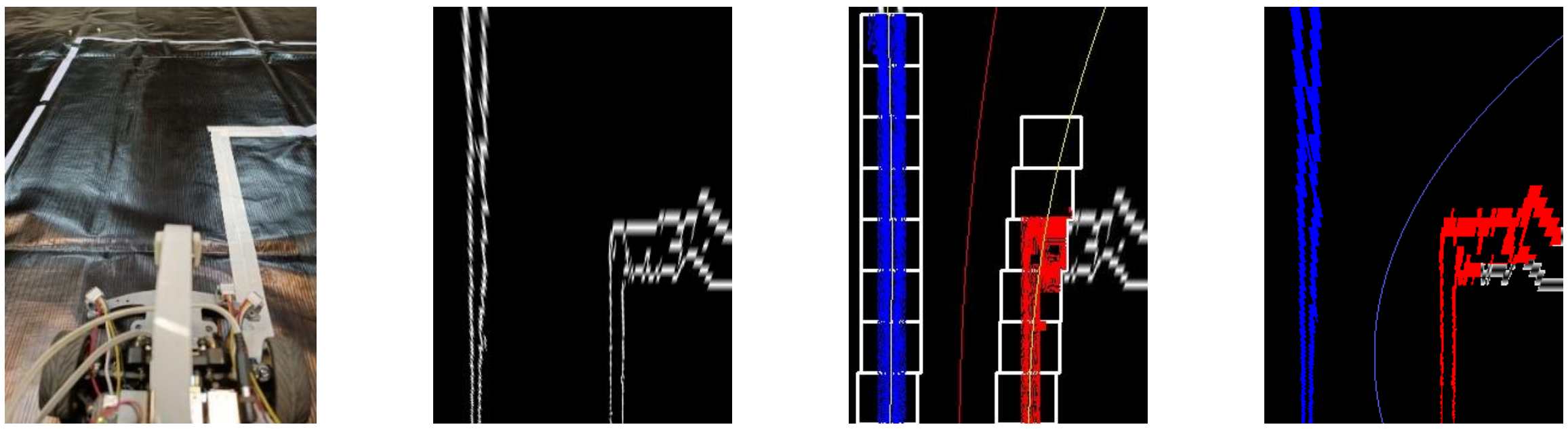

A) Position of the robot B) The binary bird's view C) Traditional sliding windows D) Elliptical searcher

Fig. 7. Comparing discussed ideal lane detection algorithms through sharp turns.

### 3.2. Controller

Figure 8 demonstrates the robot's distance from the midline over 50 seconds. The blue and yellow highlights show the system's response to road curves, and the green highlight shows the straight road after the first curve. The first discussed system used the "distance error" as the only control loop data, and the system's response is illustrated in Figure 8-A.

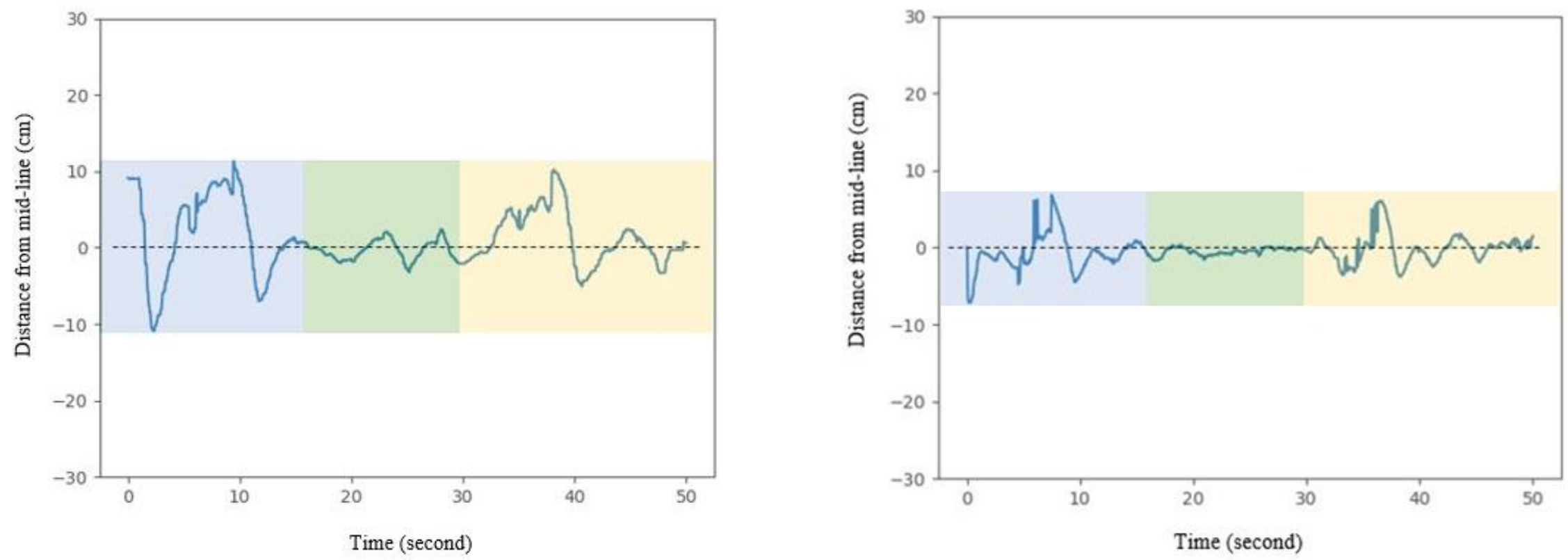

A) Using "distance error" and its derivative to control the robot

B) Using "distance error" and "angle error" along with the IMU

Fig. 8. Robot's distance from the midline in a scenario

Figure 8-B shows that the final proposed controller offers a faster, more stable, and durable response. Moreover, the suggested system offers a closer response to the midline. Yet, changing any error coefficients cannot resolve some unwanted fluctuations.

### 3.3. Sub-applications and additional remarks

#### 3.3.1. Object detection and dataset augmentation

Besides maneuvering on the road, an autonomous vehicle must be aware of traffic signs, given the current state of transportation in urban areas. To address traffic sign detection, the YOLO deep learning architecture has been practiced in this study. As a single-stage detector, YOLOv8 brings high inference speed, proper performance, and adequate learning capability. However, the lack of supervised datasets is the main barrier of object detection. To emphasize the involvement of proper data in reducing the effects of image variations, class imbalance, domain shift, and overfitting, a data augmentation and generation system is utilized, which helps train a robust model to lighting and geometrical variations. The objects of interest are augmented and pasted onto many background images, similar to the CUTMIX [11] method. Applied augmentation methods include horizontal flip, rotation, shearing, brightness and contrast alterations, saturation and hue alterations, noise injection, and random erasing with noise-injected average color filling. The set of objects included 50 arbitrarily chosen traffic signs and traffic light images. Finally, a model was trained on roughly 50K samples.

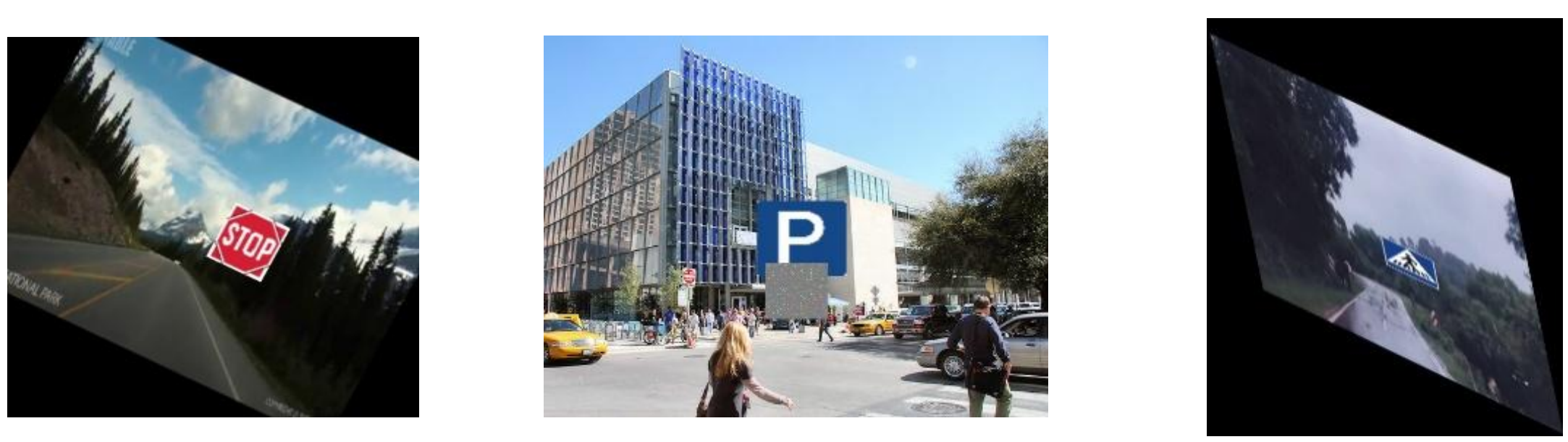

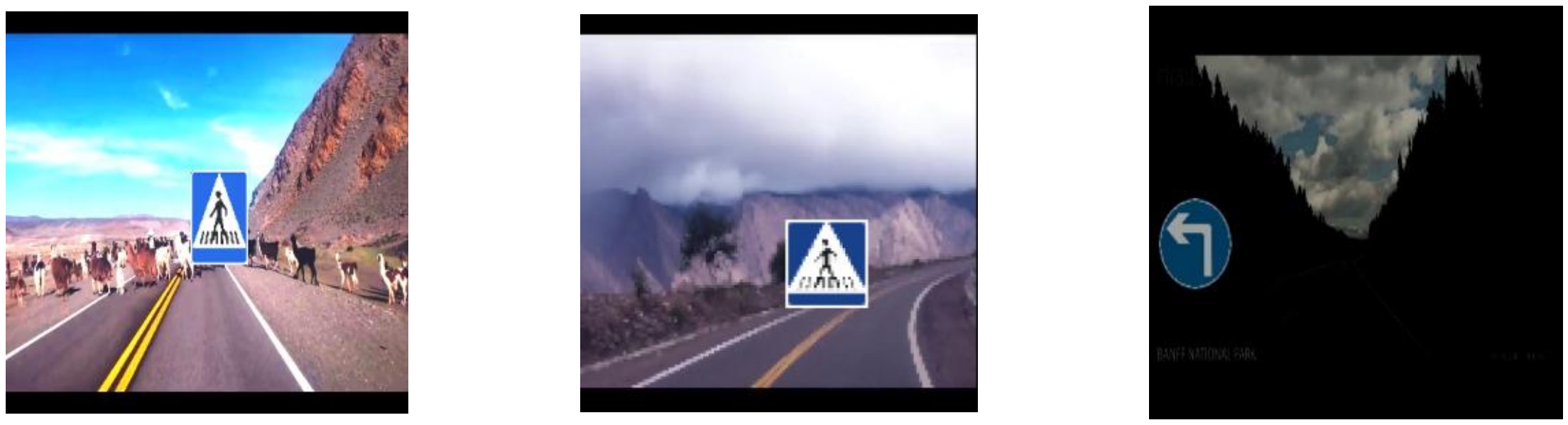

Fig. 9. Dataset image augmentation samples.

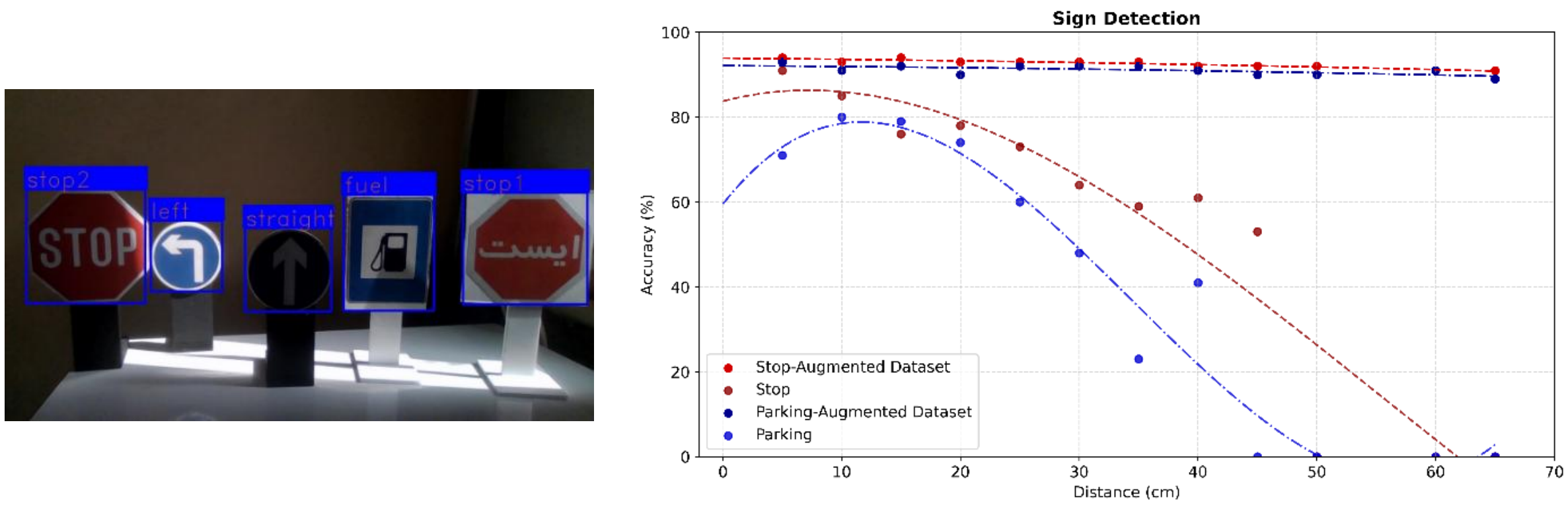

A) 5 Samples of traffic signs detection B) Distance prediction of augmented and non-augmented datasets

Fig. 10. Results of traffic sign detection

### 3.3.2. Parking sequence

A self-driving vehicle can park autonomously. As a case study and evidence of effective integration, a working parking procedure involving the detection of the Parking sign, space detection, and getting in/out of a parking space has been suggested and implemented.

Distance estimation is required for this task, assuming that the detection of the parking sign initializes the parking procedure. Single-camera distance estimation was implemented in this experiment by performing curve fitting on measurements of detected sign heights at different distances. The curve function inverse resulted in accurate sign distance estimation to initialize the parking procedure.

During the space detection phase, the system uses the approximate distance to the parking sign and data fusion from side sensors and the motor encoder to determine adequate space to initiate parking beyond the sign, ensuring that a minimum required area is available. The side sensors continuously analyze and map the roadside with regard to depth. To address potential noise that might cause false obstacle detections and prevent the vehicle from parking in suitable spaces, interpolation mechanisms are employed to refine the sensor data. Once the detection phase successfully identifies an appropriate parking spot, the car parks in the designated space.

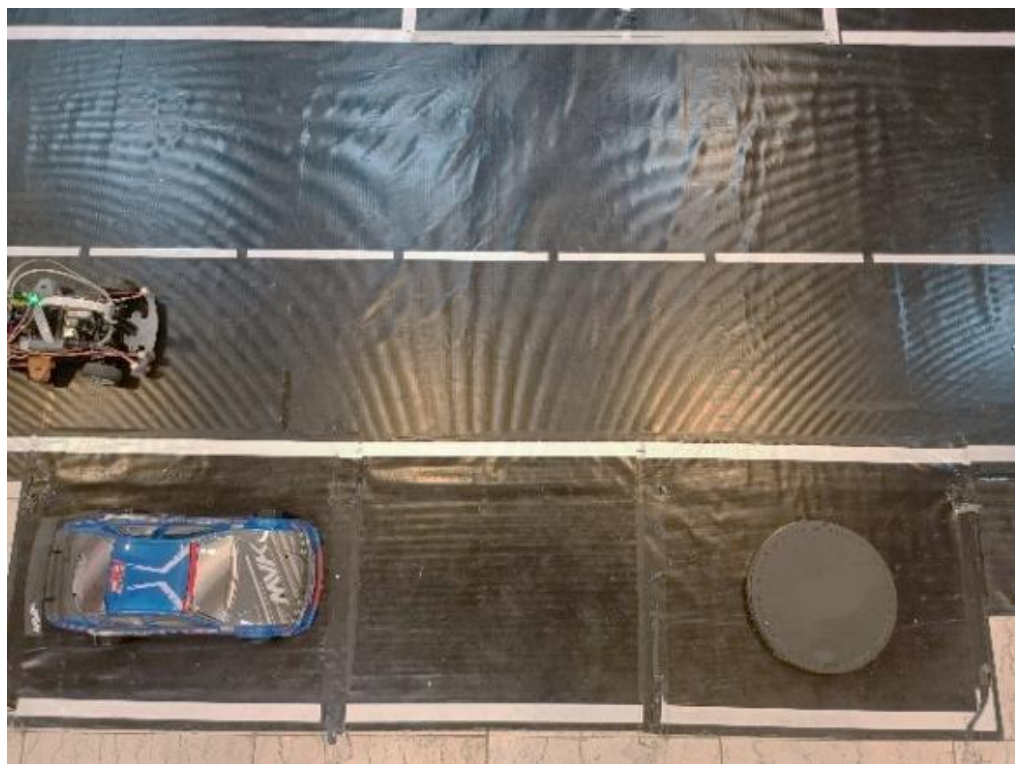

A) Position of the robot and obstacles.

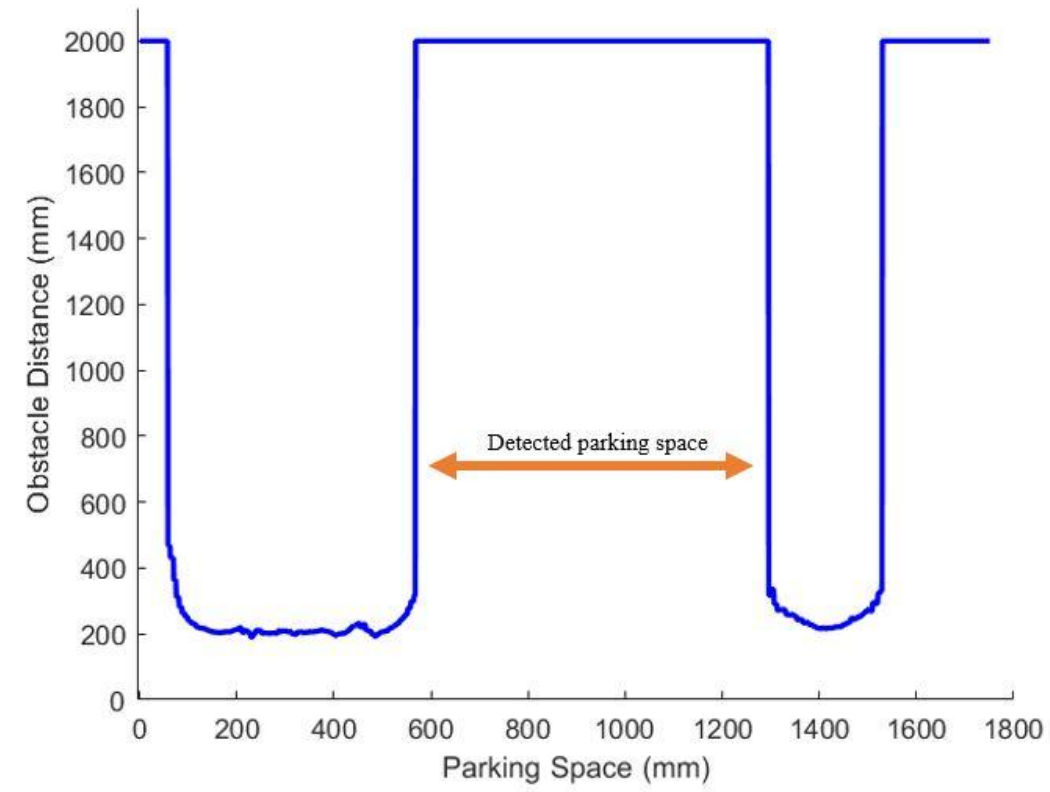

B) Result of the parking space mapping, with 2000 meaning out of range.

Fig. 11. Parking space mapping and detection

During the optimal parking procedure, the car utilizes its maximum steering angles, executing consecutive right and left turns to position itself within the parking spot accurately. After initial placement, it fine-tunes its alignment to run parallel to the road while avoiding any front or rear obstacles. When exiting the parking space, the system reassesses the space dimensions and the presence of obstacles, adjusting its path as needed without necessarily retracing its entry route. Ample available space allows for a simpler, more direct exit, whereas limited space requires additional steering maneuvers to vacate the spot safely.

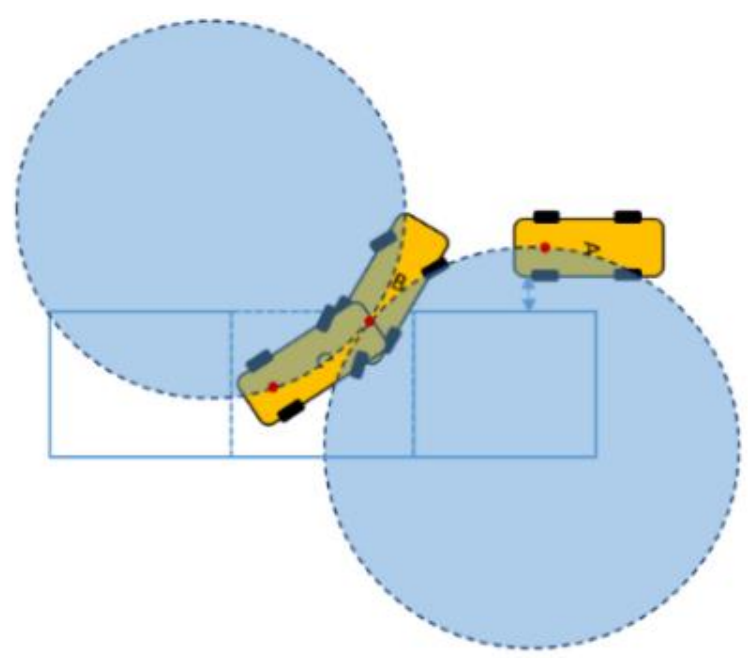

Fig. 12. Vehicle parking maneuvers

### 3.4. The platform

As described, all components of this study have been integrated to create a cost-efficient, small-scale autonomous vehicle robot. The software is managed using ROS and organized into several nodes, as shown in Figure 13. The hardware setup, illustrated in Figure 14, features the Nvidia Jetson Nano as the primary processing unit, the ESP32 as a low-cost microcontroller handling sensor fusion, control algorithms, and actuator management, along with the CMPS14 compass module and VL53L1X laser-ranging sensors.

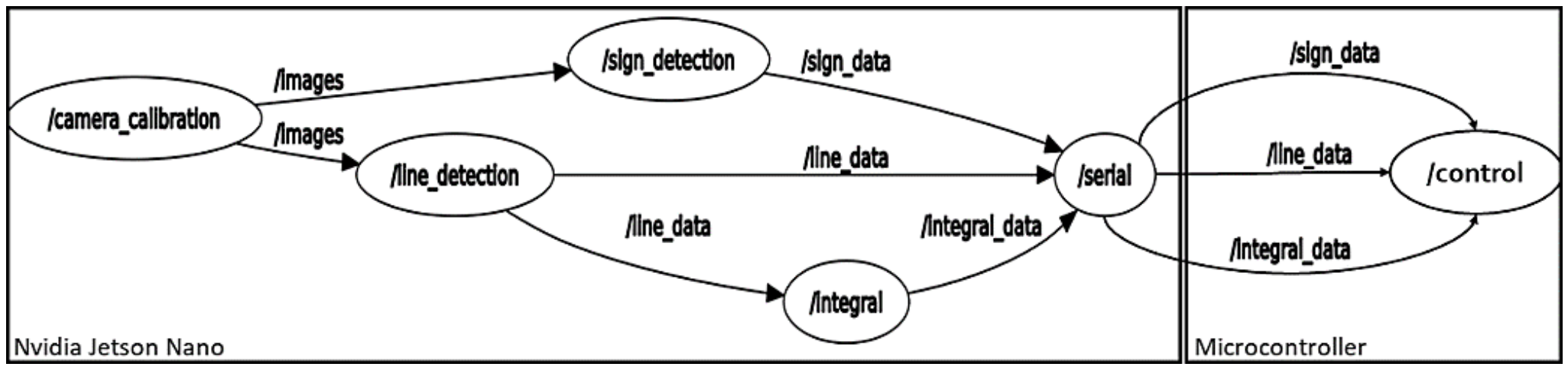

Fig. 13. ROS graph

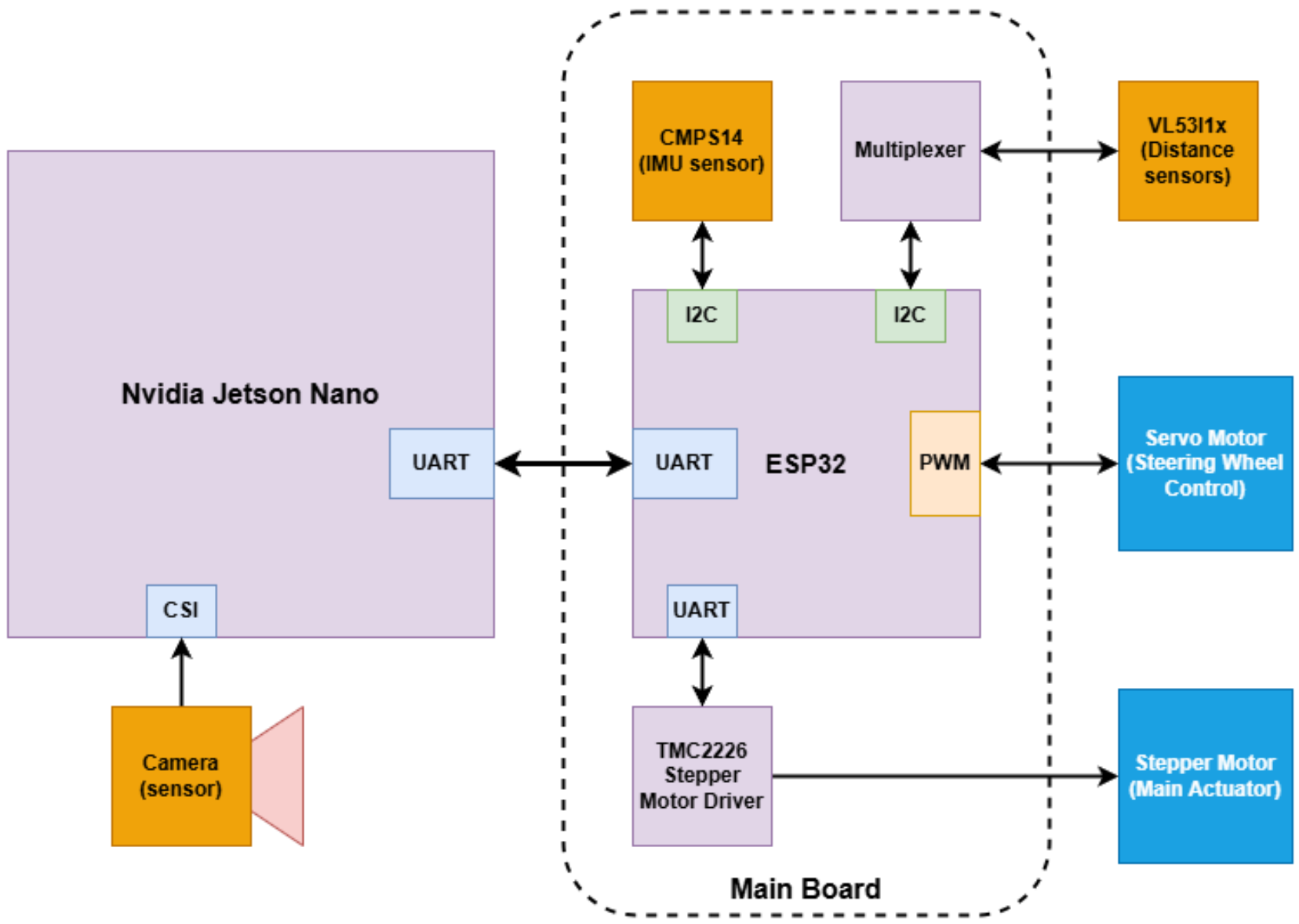

Fig. 14. Robot connections

This paper proposes the methods integrated into the stated hardware and software architectures as a platform capable of demonstrating AGV (autonomous guided vehicle) concepts and implementing future research. The authors recognize multi-robot motion planning, cooperative robotics, fleet management, SLAM, sensor fusion, image segmentation, warehouse automation, and ethical decision-making as some of the many paths of research possible to conduct with this robot platform.

## 4. Conclusion and Future Work

Autonomous transport systems, though heavily invested in, are still not reliable enough for everyday use. This paper introduces a cost-effective, small-scale autonomous vehicle platform for educational research and industrial prototyping. The platform incorporates a robust landmark tracking method for stable lane following in various lighting conditions and on curved roads. It also includes a lateral controller for smooth navigation and enhanced responsiveness to positional changes. An augmented dataset strengthens the object detection model, enabling autonomous parking. Initial testing has shown stable results, and the platform is adaptable to different configurations for research, testing, or industrial use.

The platform's modules are designed to be scalable for larger autonomous vehicles, with further development planned to enable more complex tasks like Simultaneous Localization and Mapping (SLAM) as more powerful computing systems are integrated. The final version will feature an improved exterior and mechanical design, enhancing functionality and aesthetics. This flexible, cost-efficient platform provides a solid foundation for small-scale robots and offers potential for future expansion into full-sized autonomous vehicles.